\documentclass[alpha-refs]{wiley-article}
\usepackage{siunitx}
\usepackage{multirow}
\usepackage{graphicx}
\usepackage{subcaption}

\papertype{Original Article}
\title{Reinforcement Learning with Analogical Similarity to Guide Schema Induction and Attention}

\author{James M. Foster}
\author{Matt Jones}
\affil{Department of Psychology and Neuroscience\\Institute of Cognitive Science\\University of Colorado, Boulder}
\corraddress{James M. Foster\\ Department of Psychology and Neuroscience\\ University of Colorado\\ 345 UCB\\ Boulder, CO, 80309}
\corremail{james.m.foster@colorado.edu}

\fundinginfo{This work was supported by AFOSR Grant FA-9550-10-1-0177 to Matt Jones.}

\begin{document}

\maketitle

\begin{abstract}
Research in analogical reasoning suggests that higher-order cognitive
functions such as abstract reasoning, far transfer, and creativity
are founded on recognizing structural similarities among relational
systems. Here we integrate theories of analogy with the computational
framework of reinforcement learning (RL). We propose a psychology
theory that is a computational synergy between analogy and RL, in
which analogical comparison provides the RL learning algorithm with
a measure of relational similarity, and RL provides feedback signals
that can drive analogical learning. Simulation results support the
power of this approach.

% Please include a maximum of seven keywords
\end{abstract}

\section{Introduction}

How do people learn new abstract concepts? Everyone has a repertoire
of concepts they use in making sense of the world and filtering the
infinite complexity of experience into manageable chunks. Abstract
representations that capture reoccurring patterns in the environment
are used for generalization (i.e., making predictions or inferences
about states of the environment that haven\textquoteright t been experienced
before). How are these representations constructed, and how is their
usefulness evaluated? For example, how do people discover and apply
concepts such as a 'fork' in the game of chess? How do scientists
create theories such as natural selection?

The goal of the present work is to develop a computational understanding
of how people learn abstract concepts. Previous research in analogical
reasoning suggests that higher-order cognitive functions such as abstract
reasoning, far transfer, and creativity are founded on recognizing
structural similarities among relational systems \citep{Doumas2008,Gentner1983,Hummel2003}.
However, we argue a critical element is missing from these theories,
in that their operation is essentially unsupervised, merely seeking
patterns that recur in the environment, rather than focusing on the
ones that are predictive of reward or other important outcomes.

Here we integrate theories of analogy with the computational framework
of reinforcement learning (RL). RL offers a family of learning algorithms
that have been highly successful in machine learning applications
\citep[e.g.,][]{bagnell2001,tesauro1995} and that have neurophysiological
support in the brain \citep[e.g.,][]{schultz1997}. A shortcoming
of RL is that it only learns efficiently in complex tasks if it starts
with a representation (i.e., a means for encoding stimuli or states
of the environment) that somehow captures the critical structure inherent
in the task. We formalize this notion below in terms of similarity-based
generalization \citep{shepard1987} and kernel methods from statistical
machine learning \citep{shawe-taylor2004}. In other words, RL requires
a sophisticated sense of similarity to succeed in realistically complex
tasks. Psychologically, the question of how such a similarity function
is learned can be cast as a question of learning sophisticated, abstract
representations.

This paper proposes a computational model of analogical RL, in which
analogical comparison provides the RL learning algorithm with a measure
of relational similarity, and RL provides feedback signals that can
drive analogical learning. Relational similarity enables RL to generalize
knowledge from past to current situations more efficiently, leading
to faster learning. Conversely, the prediction-error signals from
RL can be used to guide induction of new higher-order relational concepts.
Thus we propose there exists a computationally powerful synergy between
analogy and RL. The simulation experiment reported here supports this
claim. Because of the strong empirical evidence for each of these
mechanisms taken separately, we conjecture that the brain exploits
this synergy as well.

We want to emphasize that this isn't a theory of specific psychological
phenomena. We're coming from the perspective that human conceptual
learning and invention is far beyond current scientific explanation.
There are no existing models that can compare to what the brain achieves.
Our goal is to explore where this power might come from. The computational
framework we propose is grounded in well-established psychological
principles that themselves are supported by large bodies of experimental
evidence, but the aim of our model is not to explain specific data.
It's to demonstrate the potential power that comes from combining
these principles in the way we propose. Thus the scope of this paper
is to understand how the proposed mechanism might work in principle.
Future work will derive testable predictions and empirical means for
testing them.

In the following sections, we will review analogy and RL, then lay
out our computational proposal and present a formal model, then present
simulation results and discuss implications and limitations of the
model.

\section{Analogy}

Research in human conceptual knowledge representation has shown that
concepts are represented not just as distributions of features \citep[cf.][]{Nosofsky1986,rosch1975}
but as relational structures. This relational knowledge includes both
internal structure, such as the fact that a robin's wings allow it
to fly \citep{sloman1998}, as well as external structure, such as
the fact that a dog likes to chase cats \citep{Jones2007}. Theories
of analogical reasoning represent relational knowledge of this type
in a predicate calculus that binds objects to the roles of relations,
for example \textsc{chase(dog,cat)}. According to these theories,
an analogy between two complex episodes (each a network of relations
and objects) amounts to recognition that they share a common relational
structure \citep{Gentner1983,Hummel2003}.

\begin{figure}[htb]
\begin{centering}
\includegraphics[scale=0.6]{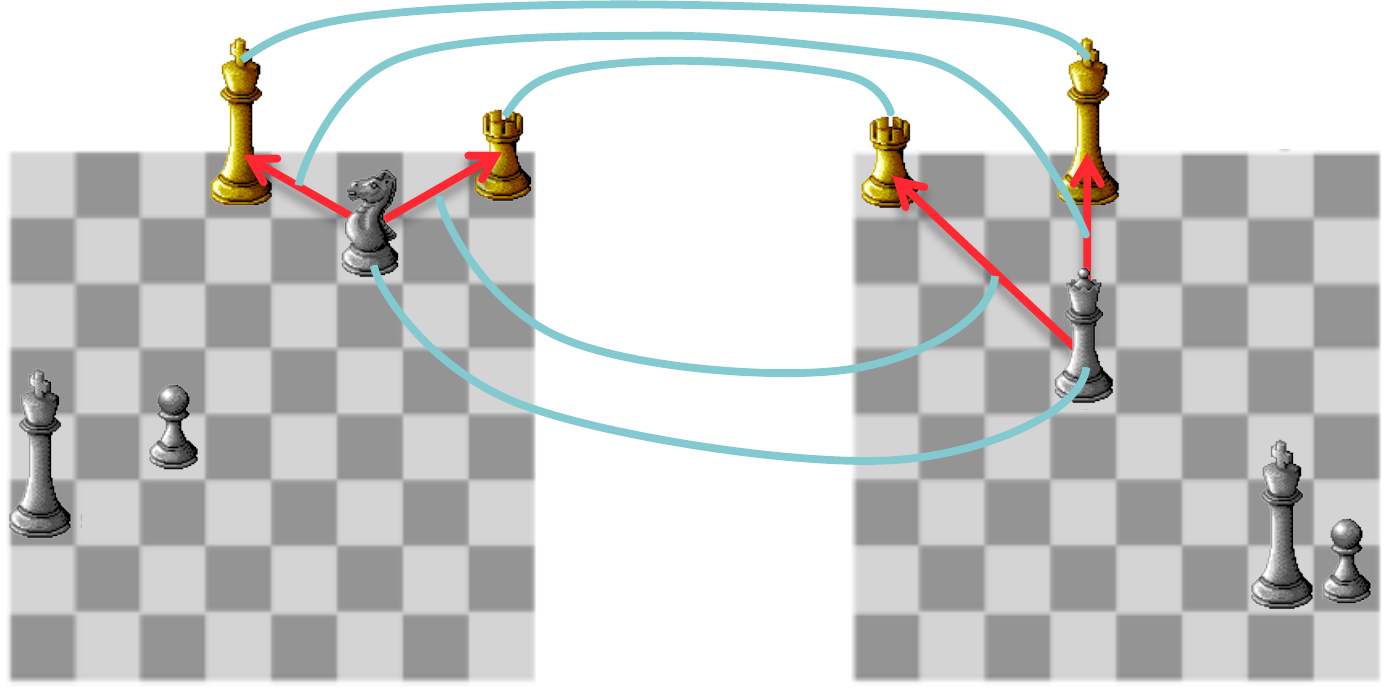} 
\par\end{centering}
\protect\caption{An example of structural alignment between two chess positions. Both
positions contain instances of the abstract concept of a \textsc{fork}:
black's piece is simultaneously attacking both of white's pieces.
These attacking relations are represented by the red arrows. Cyan
lines indicate the mapping between the two episodes. The mapping satisfies
parallel connectivity because it respects the bindings between relations
and their role-fillers.}
\label{fig:chess-1} 
\end{figure}

At a more mechanistic level, the dominant theory of analogy is \textit{structural
alignment} \citep{Gentner1983}. This process involves building a
mapping between two episodes, mapping objects to objects and relations
to relations. The best mapping is one that maps objects to similar
objects, maps relations to similar relations, and most importantly,
satisfies \textit{parallel connectivity}. Parallel connectivity means
that, whenever two relations are mapped to each other, the objects
filling their respective role-fillers are also mapped together. An
example is shown in Figure \ref{fig:chess-1}. Parallel connectivity is satisfied here
because, for each mapped pair of \textsc{attack} relations (red arrows),
the objects filling the \textsc{attacker} role are mapped together
(knight is mapped to queen), and the objects filling the \textsc{attacked}
role are also mapped together (rook to rook and king to king). Thus
structural alignment constitutes a (potentially partial or imperfect)
isomorphism between two episodes, which respects the relational structure
that they have in common. Importantly, if the search for a mapping
gives little emphasis to object-level similarity (as opposed to relation-level
similarity and parallel connectivity), then structural alignment can
find abstract commonalities between episodes having little or no surface
similarity (i.e., in terms of perceptual features).

%h here, t top, b bottom

We propose structural alignment is critical to learning of abstract
concepts for three reasons. First, perceived similarity of relational
stimuli depends on structural alignability \citep{markman1993}. Second,
structural alignment is important for analogical transfer, which is
the ability to apply knowledge from one situation to another, superficially
different situation \citep{gick1980}. For example, a winning move
in one chess position can be used to discover a winning move in a
different (but aligned) position, by translating that action through
the analogical mapping. Third, a successful analogy can lead to \textit{schema
induction}, which involves extraction of the shared relational structure
identified by the analogy \citep{Doumas2008,Gentner1983,Hummel2003}.
In the example of Figure \ref{fig:chess-1}, this schema would be a system of relational
knowledge on abstract (token) objects, including \textsc{attack(piece1,piece2)},
\textsc{attack(piece1,piece3)}, and potentially other shared information
such as \textsc{not\_attacked(piece1)} and \textsc{king(piece2)}.

These three observations suggest that analogy plays an important role
in learning and use of abstract relational concepts. The first two
observations suggest that analogical transfer is like similarity-based
generalization, but it's also more sophisticated because it takes
structure into account, as we elaborate in the next two sections.
In brief, structural alignment offers a sophisticated form of similarity
that can be used to generalize knowledge between situations that are
superficially very different. The third observation suggests that
analogy can discover new relational concepts (e.g., the concept of
a chess fork, from Figure \ref{fig:chess-1}), which can in turn lead to perception
of even more abstract similarities among future experiences.

One potential shortcoming of the basic theory of analogy reviewed
here is that is it essentially unsupervised. In this framework, the
quality of an analogy depends only on how well the two systems can
be structurally aligned, and not on how useful or predictive the shared
structure might be. For example, one could list many relational patterns
that arise in chess games but that are not especially useful for choosing
a move or for predicting the course of the game. In previous work,
we have found that implementing structural alignment and schema induction
in a rich and structured artificial environment results in discovery
of many frequent but mostly useless schemas \citep{foster2012}. An
alternative, potentially more powerful model of analogical learning
would involve feedback from the environment, so that the value of
an analogy or schema is judged partially by how well it improves predictions
of reward or other important environmental variables. For example,
the concept of a fork in chess is an important schema not (only) because
it is a recurring pattern in chess environments, but because it carries
information about significant outcomes (i.e., about sudden changes
in each player's chances of winning). A natural framework for introducing
this sort of reward sensitivity into theories of analogy is that of
RL, which we review next.

\section{Reinforcement Learning}

RL is a mathematical and computational theory of learning from reward
in dynamic environments. An RL task is characterized by an agent embedded
in an environment that exists in some \textit{state} at any given
moment in time. At each time step, the agent senses the state of its
environment, takes an action that affects what state occurs next,
and receives a continuous-valued reward that depends on the state
and its action \citep{sutton1998}. This framework is very general
and can encompass nearly any psychological task in which the subject
has full knowledge of the state of the world at all times (i.e., there
are no relevant hidden variables).

Most RL models work by learning \textit{values} for different states
or actions, which represent the total future reward that can be expected
from any given starting point (i.e., from any state or from any action
within a state). These values can be learned incrementally, from \textit{temporal-difference
(TD) error} signals calculated from the reward and state following
each action (see Model section). There is strong evidence that the
brain computes something close to TD error, and thus that RL captures
a core principle of biological learning \citep{schultz1997}.

In principle, this type of simple algorithm could be used to perfectly
learn a complex task such as chess, by experiencing enough games to
learn the true state values (i.e., probability of winning from every
board position) and then playing according to those values. However,
a serious shortcoming of this naive approach is that it learns the
value of each state independently, which can be hopelessly inefficient
for realistic tasks that typically have very large state spaces. Instead,
some form of generalization is needed, to allow value estimates for
one state to draw on experience in other, similar states.

Many variants of RL have been proposed for implementing generalization
among states \citep[e.g.,][]{albus1981,sutton1988}. Here we pursue
a direct and psychologically motivated form of generalization, based
on similarity \citep{jones2010,ormoneit2002}. We assume the model
has a stored collection of exemplar states, each associated with a
learned value. This exemplar representation is particularly suited
for the analogy model we present below because it allows us to treat
schemas as exemplars. The value estimate for any state is obtained
by a similarity weighted average over the exemplars' values; that
is, knowledge from each exemplar is used in proportion to how similar
it is to the current state. This approach is closely related to exemplar-generalization
models in more traditional psychological tasks such as category learning
\citep{Nosofsky1986}. It can also be viewed as a subset of kernel
methods from machine learning \citep{shawe-taylor2004}, under the
identification of the kernel function with psychological similarity
\citep{jakel2008}.

A critical consideration for all learning models (including RL models)
is how well their pattern of generalization matches the inherent structure
of the task. If generalization is strong only between stimuli or states
that have similar values or outcomes, then learning will be efficient.
On the other hand, if the model generalizes significantly between
stimuli or states with very different outcomes, its estimates or predictions
will be biased and learning and performance will be poor. The kernel
or exemplar-similarity approach makes this connection explicit, because
generalization between two states is directly determined by their
similarity. As we propose next, analogy and schema induction offer
a sophisticated form of similarity that is potentially quite powerful
for learning complex tasks with structured stimuli.

\section{Analogical RL}

The previous two sections suggest a complementary relationship between
analogy and RL, which hint at the potential for a computationally
powerful, synergistic interaction between these two cognitive processes.
We outline here a formal theory of this interaction. The next two
sections provide a mathematical specification of a partial implementation
of this theory, and then present simulation results offering a proof-in-principle
of the computational power of this approach.

The first proposed connection between analogy and RL is that structural
alignment yields an abstract form of psychological similarity that
can support sophisticated generalization \citep{gick1980,markman1993}.
Incorporating analogical similarity into the RL framework could thus
lead to rapid learning in complex, structured environments. For example,
an RL model of chess equipped with analogical similarity and a notion
of an \textit{attack} relation should recognize the similarity between
the two positions in Figure \ref{fig:chess-1} and hence generalize between them. Consequently
the model should learn to create forks and to avoid forks by the opponent
much more rapidly than if it had to learn about each possible fork
instance individually.

The second proposed connection is that the TD error computed by RL
models, for updating value estimates, can potentially drive analogical
learning by guiding schema induction and attention. Instead of forming
schemas for whatever relational structures are frequently encountered
(or are discovered by analogical comparison of any two states), an
analogical RL model can be more selective, only inducing schemas from
analogies that significantly improve reward prediction. Such analogies
indicate that the structure common to the two analogue states may
have particular predictive value in the current task, and hence that
it might be worth extracting as a standalone concept. For example,
if the model found a winning fork move by analogical comparison to
a previously seen state involving a fork, the large boost in reward
could trigger induction of a schema embodying the abstract concept
of a fork. Furthermore, the model can use prediction error to increase
attention to concepts that are more useful for predicting reward,
and decrease attention to concepts that are less useful. Such an attention-learning
mechanism is proposed to bias the model to rely more on concepts that
have been consistently useful in the past.

The proposed model thus works as follows (see the next section for
technical details). The model maintains a set of exemplars $E$, each
with a learned value, $v(E)$. To estimate the value of any state
$S$, it compares that state to all exemplars by structural alignment,
which yields a measure of analogical similarity for each exemplar
\citep{forbus1989}. The estimated value of the state, $\tilde{V}(s)$,
is then obtained as a similarity-weighted average of $v(E)$. After
any action is taken and the immediate reward and next state are observed,
a TD error is computed as in standard RL. The exemplar values are
then updated in proportion to the TD error and in proportion to how
much each contributed to the model's prediction, that is, in proportion
to $\mathrm{sim}(s,E)$. 

The attentions $u(E)$ implement exemplar-specific attentions weights
or learning rates. Attention learning is an additional mechanism of
the model whose purpose is to increase the model's reliance on useful
exemplars. Although the model makes sense without this attention learning
mechanism, including it improves performance and integrates analogy,
RL, and attention, and has been demonstrated in experiments with humans \citep{foster2013schemas,foster2015analogical}.
A model that increases its repertoire of concepts needs some
pruning mechanism to sort through what's been discovered. This attention learning approach is a reasonable way to handle the pruning problem in an exemplar setting: the model
needs to learn which exemplars or schemas to retain, so we attach
an attention value to each one.  

The $u(E)$ values can also be thought of as voting weights in the
computation of $\tilde{V}(s)$. Exemplars with higher $u$ values
have greater influence on the similarity-weighted average of exemplar
values. The attentions are updated in proportion to the TD error,
in proportion to how much each contributed to the model's prediction,
and in proportion to how much that exemplar's prediction differed
from the overall prediction $\tilde{V}$. Attention is increased to
exemplars that individually made a more accurate prediction that the
overall prediction, and attention is decreased to exemplars that individually
made a less accurate prediction than the overall prediction.

Analogy and RL also mutually facilitate each other through an additional
mechanism of schema induction. Whenever the structural alignment between
a state and an exemplar produces a sufficient reduction in prediction
error (relative to what would be expected if that exemplar were absent),
a schema is induced from that analogy. The schema is an abstract representation,
defined on token (placeholder) objects, and it contains only the shared
information that was successfully mapped by the analogy. The schema
is added to the pool of exemplars, where it can acquire value associations
directly (just like the exemplars do). Theoretically, we take the
position that there is no real psychological difference between schemas
and concrete state exemplars, it's just a continuum of specificity. The advantage
conferred by the new schema is that it allows for even faster learning
about all states it applies to (i.e., that contain that substructure).
For example, rather than learning by generalization among different
instances of forks, the model would learn a direct value for the fork
concept, which it could immediately apply to any future instances.
A consequence of the schema induction mechanism is that the pool of exemplars comes to contain more and more abstract schemas.
Thus the model's representation transitions from initially episodic
to more abstract and conceptual. The facilitation between analogy
and RL here is that analogy provides new representations for RL to
use for its value learning, and RL provides a TD error signal to help
analogy decide which schemas to build. 

Analogical RL thus integrates three principles from prior research:
RL, exemplar generalization, and structural alignment of relational
representations. Because each of these principles has strong empirical
support as a psychological mechanism, it is plausible that they all
interact in a manner similar to what we propose here. Thus it seems
fruitful to explore computationally what these mechanisms can achieve
when combined.

\section{Model}

\begin{figure}[ht]
\includegraphics[width=1\textwidth]{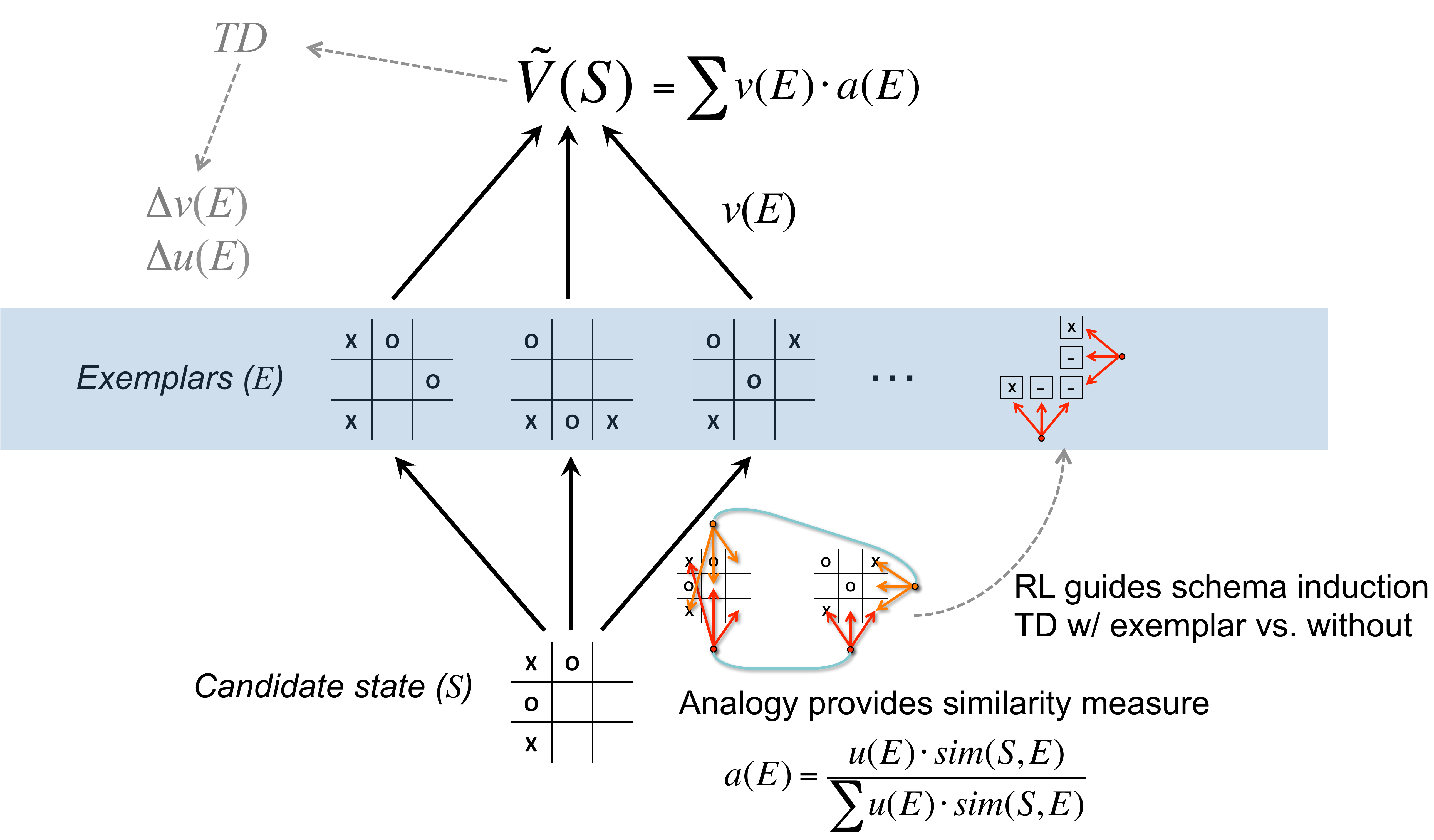}
\protect\caption{Schema Induction Model operation. Each candidate afterstate
is evaluated by analogical comparison to stored exemplars, followed
by similarity-weighted averaging among the learned exemplar values.
Learning is by TD error applied to the exemplar values. On some trials,
especially useful analogies produce new schemas that are added to
the exemplar pool. In the example here, $S$ and $E$ both have guaranteed
wins for X by threatening a win in two ways. The induced schema embodies
this abstract structure. Dots with red arrows indicate ternary ``same-rank''
relations.}
\label{fig:SchemaInductionModel}
\end{figure}

The proposed model applies to Markov Decision Process tasks, where
an agent makes decisions based on the current state of the environment.
At each time step, the agent chooses from the available actions in
the current state and then the environment gives the agent an immediate
reward and moves into the next state. 

The simulation study presented below uses a variant of RL known as
\textit{afterstate learning}, in which the agent learns values for
the possible states it can move into \citep{sutton1998}. This is
a reasonable and efficient method for the task we use here---tic-tac-toe,
or noughts \& crosses---because the agent's opponent can be treated
as part of the environment and is the only source of randomness. Our
main proposal regarding the interaction between RL and analogical
learning is not limited to this approach.

The operation of the model is illustrated in Figure~\ref{fig:SchemaInductionModel}.
On each time step, the model identifies all possible actions and their
associated afterstates. For each afterstate $S$, it computes an analogical
similarity, $sim(S,E)$, to each exemplar, $E$, by structural alignment.
At the theoretical level, the model does not commit to a particular
mapping algorithm. Instead, we assume the mapping could be done by
any of the extant mapping models \citep[e.g.,][]{Falkenhainer1989,Larkey2003,Hummel2003,hofstadter1994copycat}.
Each possible mapping $M:S\rightarrow E$ is evaluated according to
\begin{equation}
\Phi(M)=\beta\cdot\sum_{o\in s}\mathrm{sim}\left(o,M(o)\right)+\sum_{r\in s}\mathrm{sim}\left(r,M(r)\right)\cdot\left[1+\sum_{i=1}^{n_{r}}I_{\left\{ M\left(\mathrm{child}_{i}(r)\right)=\mathrm{child}_{i}\left(M(r)\right)\right\} }\right].
\end{equation}
This expression takes into account object similarity, by comparing
each object $o$ in $S$ to its image in $E$; relational similarity,
by comparing each relation $r$ in $S$ to its image in $E$; and
parallel connectivity, by having similarity between mutually mapped
relations ``trickle down'' to add to the similarity of any mutually
mapped role-fillers \citep{forbus1989}. The $\mathrm{sim}$ function
is a primitive (object- and relation-level) similarity function, $\beta$
determines the relative contribution of object similarity, $n_{r}$
is the number of roles in relation $r$, $\mathrm{child}_{i}(r)$
is the object filling the $i^{\mathrm{th}}$ role of $r$, and $I_{\{P\}}$
is an indicator function equal to $1$ when proposition $P$ is true.

A more intuitive understanding of the $\Phi(M)$ equation may come
from a description of its implementation, which works as follows. Given
a mapping, a schema is created for that mapping (one node for each
pair of mapped nodes, with links between schema nodes when there's
parallel links in $S$ and $E$. The schema nodes are sorted by concept
type, so that higher-order concepts are processed first. Then, each
node is processed by incrementing its score by a constant, and incrementing
its child nodes' scores by a trickle down factor times its score.
Finally, each node's final score is summed to get a total score for
the mapping. This trickle down scoring gives a bonus to systematic
mappings (those that share systems of relations governed by common
higher-order relations). 

Analogical similarity is then defined as the value of the best mapping
(here the $\theta$ parameter determines specificity of generalization):
\begin{equation}
\noindent 
\label{eq:analogical_similarity}
sim(S,E)=\exp\left(\theta\cdot\max_{M}\Phi(M)\right).
\end{equation}
%need to be consistent with sim(S,E) vs. K(s,E)
The activation $a(E)$ of each exemplar is determined by weighting
the analogical similarity between that exemplar \textit{E} and the
candidate state \textit{S} by its attention \textit{u} and normalizing
by the attention-weighted analogical similarity across all exemplars:
\begin{equation}
a(E)=\frac{u(E)\cdot sim(S,E)}{\underset{E'\in Exemplars}{\sum}u(E')\cdot sim(S,E')}.
\end{equation}
The estimated value of $S$, $\tilde{V}(S)$, is computed as a similarity-weighted
average of the exemplar values $v(E)$:
\begin{equation}
\tilde{V}(S)=\sum v(E)\cdot a(E).
\end{equation}
Thus the estimate is based on the learned values of the exemplars
most similar to the candidate state and the exemplars with the highest
attention weights. 

There is a separate pass through the whole network for each candidate
state (i.e., an outer loop over candidates). Although our model evaluates
all possible afterstates, a more realistic model would be selective.
How an agent decides which options to even consider is an important
question but is outside the scope of this paper.

Once values $\tilde{V}(S)$ have been estimated for all candidate
afterstates, the model uses a softmax (Luce-choice or Gibbs-sampling
rule) to select what state to move into (here $\tau$ is an exploration
parameter):
\begin{equation}
\Pr[S_{t}=S]\propto e^{\tilde{V}(S)/\tau}.
\end{equation}

Learning based on the chosen afterstate $S_{t}$ follows the SARSA
rule \citep{rummery1994}, after the model chooses its action on the
next time step. This produces a TD error, which is then used to update
the exemplar values and exemplar attention or voting weights by gradient
descent. Exemplar values are updated in proportion to the TD error
and their activations:
\begin{equation}
\triangle v(E)=\epsilon\cdot TD\cdot a(E)
\end{equation}
where $\epsilon$ is a learning rate. Exemplar attentions are updated
in proportion to the learning rate $\epsilon$, the TD error, the
analogical similarity between the exemplar and candidate state $sim(S,E)$,
and the difference between the exemplar value $v(E)$ and the estimated
value of the candidate state $\tilde{V}(S)$, normalized by the attention-weighted
analogical similarity across all exemplars:
\begin{equation}
\Delta u(E)=\frac{\epsilon\cdot TD\cdot sim(S,E)\cdot(v(E)-\tilde{V}(S))}{\underset{E'\in Exemplars}{\sum}u(E')\cdot sim(S,E')}\;.\label{eq:deltaU}
\end{equation}

This exemplar-specific attention weight learning mechanism (\emph{u}
parameter in the equations) allows the model to learn which exemplars
are most useful for reducing prediction error. Over time, the model
learns to increase attention to exemplars which are predictive of
reward outcome, and decrease attention to exemplars which are not
predictive of outcome. 

Following learning after each trial, the schema induction mechanism
determines how much each exemplar contributed to reducing prediction
error, by comparing TD to what it would have been without that
exemplar. 
%TODO: this equation needs completing
%\begin{equation}
%TD_{reduction}(E)= r(t)+\gamma\cdot\tilde{V}(S)...
%\end{equation}
If the reduction is above some threshold, the analogical
mapping found for that exemplar (lower right of Figure \ref{fig:SchemaInductionModel})
produces a schema that is added to the exemplar pool (far right).
The schema is given a value of $v$ initialized at $\tilde{V}(S_{t})$.
This schema value is updated on future trials just as are the exemplar
values. Acquisition of new schemas in this way is predicted to improve
the model's pattern of generalization, tuning it to the most useful
relational structures in a task.\footnote{Schemas can be spawned from the mapping between a
candidate state and an exemplar state (\textit{schema induction}) as well as from the mapping between a candidate state and an exemplar schema \citep[\textit{schema refinement;}][]{Doumas2008}}

\section{Simulation}

The goal of the simulation was to test whether the model could (1)
learn more quickly by generalizing between states sharing structure
but not surface similarity and (2) bootstrap its learning by discovering
composite structures (schemas) that are particularly predictive. If
the model succeeded at these two measures, then that would suggest
it has potential to exhibit humanlike behavior in more complex settings.
For example, the model might offer insight into how humans become
experts in complex games like chess and Go, or how they acquire and
apply relationally complex sentence structures \citep[see][]{goldwater2011structural}.

The analogical RL model was tested on its ability to learn tic-tac-toe
\citep{foster2013}. Tic-tac-toe was chosen as a test domain because
it is a simple game with relational structure and a clear task goal.
However, we want to be clear that this is not a theory of how people
play tic-tac-toe. People are typically given explicit instruction
on the rules, goals, and winning states of the game. The model isn't
provided any rules of the game, other than implicitly knowing what
moves are legal at any point (i.e., you can only move in an empty
square) and the model doesn't know what constitutes a win. Instead,
the model learns entirely from trial and error. Whereas people are
able to look ahead and mentally simulate moves several steps into
the future, the model as currently implemented does not look ahead
beyond the current move. 

\begin{table}[b]
\caption{Model Variations.}
\label{table:ModelVariations}
\begin{threeparttable}
\begin{tabular}{p{6cm} p{6cm}}
\headrow
\thead{Model Variation} & \thead{Additional Mechanism Tested}\\
Featural & \\
Relational &  Relational similarity\\
Unguided Schema Induction, Fixed Attentions & Schema induction\\
Guided Schema Induction, Fixed Attentions & Schema induction guided by RL\\
Guided Schema Induction, Learned Attentions & State and Schema attentions guided by RL\\
\hiderowcolors
\hline  % Please only put a hline at the end of the table
\end{tabular}
%\begin{tablenotes}
%\item JKL, just keep laughing; MN, merry noise.
%\end{tablenotes}
\end{threeparttable}
\end{table}

\begin{figure}[ht]
\centering
\includegraphics[width=.5\textwidth]{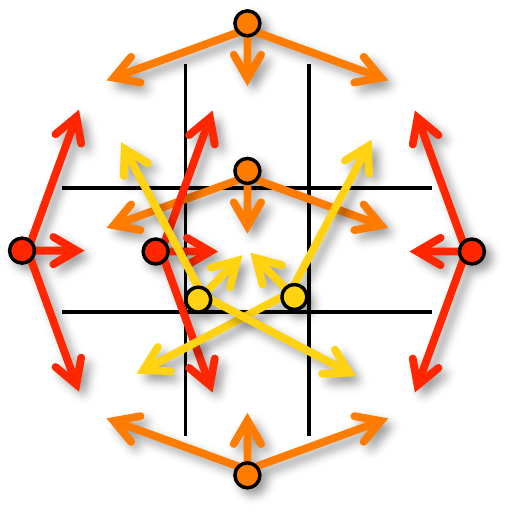}
\protect\caption{Relations Defined on the Tic-Tac-Toe board. The figure illustrates the 8 ternary relations that the Relational Model uses to represent a state of the game. There are three relations for the 3 vertical columns (in red), 3 relations for the 3 horizontal rows (in orange), and 2 relations for the diagonals (in yellow). Each relation has 3 roles, which are filled by the objects in their corresponding board locations.}
\label{fig:RelationsInTTT}
\end{figure}

In the simulation, each board position was represented by treating
the nine squares as objects of types 0 (blank), 1 (focal agent's),
and 2 (opponent's), and defining 8 ternary ``same-rank'' relations
for the rows, columns, and diagonals. Thus a player wins by filling
all squares in any one of these relations (see Figure \ref{fig:RelationsInTTT}).
Object similarity was defined as 1 for matching object types and 0
otherwise. Similarity between relations was always 1 because there
was only one type of relation. Reward was given only at the end of
a game, as +1 for the winner, -1 for the loser, or 0 for a draw. After
the game ended, it moved to a special terminal state with fixed value
of 0. For simplicity, all free parameters of the model ($\beta,\theta,\alpha,\gamma,\tau$)
were set to a default value of 1, except the schema induction threshold which was set to 6 standard deviations above the mean TD error reduction on each turn. 
%This threshold limited schemas to be induced only when the reduction in TD error by an exemplar was 6 standard deviations above the mean reduction in TD error by all the exemplars. 
% r26r27 parameters: recruitment4 SI6 fixedyoking1.25 LR1

\begin{figure}[ht]
\centering
\includegraphics[width=1\textwidth]{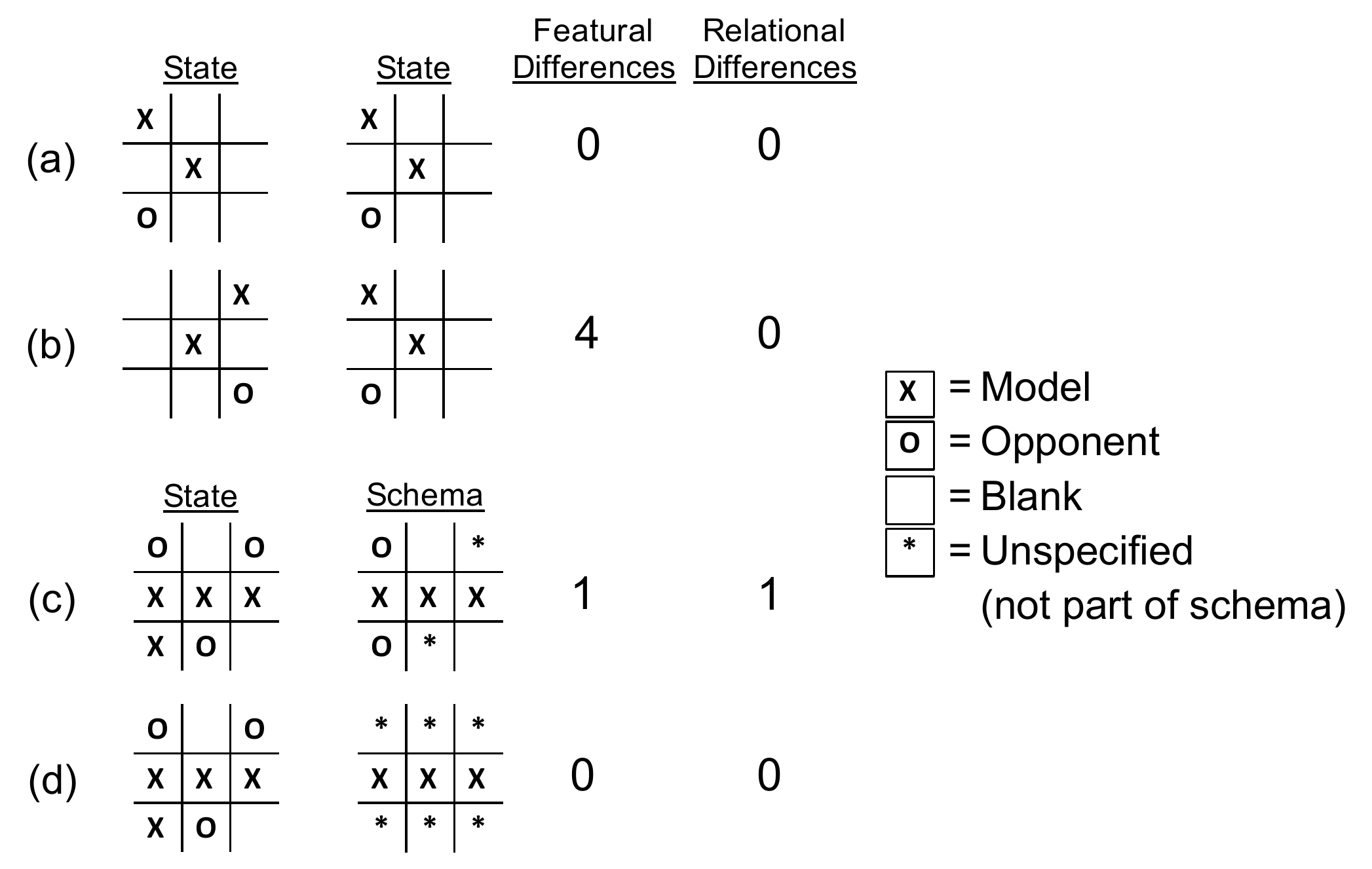}
\protect\caption{Featural vs. Relational Similarity in the Tic-Tac-Toe implementation. X indicates the model's token, O indicates the opponent's token, blank indicates an open space on the grid. Similarity is inversely related to the number of differences between states and states, or between states and schemas. Asterisks indicate board locations that are not part of the schema, and so do not contribute to the count of differences. (a) The featural and relational model similarity would be the same because all tokens match based on their absolute board locations. (b)  The relational model would score these two states as having higher similarity because it allows for the reflection of the board, whereas the featural model would count four differences. (c) The schema on the right contains three X's in a row, as well as two O's, two blanks, and two locations that are not part of the schema. (d) The schema on the right has been further refined, and is "clean" in that it only represents the information relevant to a winning board configuration, 3 X's in a row}
\label{fig:featural_vs_relational_sim}
\end{figure}

State exemplars (i.e., non-schema exemplars) were added to the model probabilistically, with probability
of recruitment inversely proportional to the number of state
exemplars already recruited. Recruitment of duplicate exemplars
was not allowed. 

Five variations of the model were implemented to verify its mechanisms.
Each model variation builds on the previous variation by adding an
additional mechanism (see Table \ref{table:ModelVariations}). 

The Featural model was restricted to literal mappings between states
(upper-left square to upper-left square, etc.). This model still included
generalization, but its similarity was restricted to the concrete
similarity of standard feature-based models. Featural similarity was
defined as the proportion of matching objects in the same absolute
board locations.  

The Relational model considered all 8 mappings defined by rigid rotation
and reflection of the board. This scheme was used in place of searching
all $9!$ possible mappings for every comparison, to reduce computation
time (see Figure \ref{fig:featural_vs_relational_sim}). 

The Unguided Schema Schema Induction, Fixed Attentions model extended
the Relational model by inducing schemas that capture the relational
structure critical to the task. This model variation included schema
induction and relational similarity, but the schema induction process
was not guided by RL feedback. Instead, this baseline model randomly
induced schemas between exemplars and the current state (or between
schemas and the current state). The number of schemas induced was
determined by yoking to the feedback-guided schema induction model.
Thus this yoked baseline model induced the same number of schemas
as the feedback-guided schema induction model, but the particular
schemas learned were induced from comparisons between the current
state and randomly chosen exemplars. The hypothesis was that the feedback-guided
schema induction model would learn faster than the unguided model,
and would also discover more useful representations.
% * <jmfoster@gmail.com> 2017-12-23T23:13:58.813Z:
%
% random induction and yoking could be explained more.  at each ply, the chosen move was mapped to all items in the exemplar pool.  every such mapping is a candidate for schema induction.  some number of these were randomly selected and the resulting schemas added to the pool.  the number selected was determined by the number of schemas induced by the full model -- on that exact trial number, or somehow smoothed over blocks?  incidentally, it might help readers to have some idea of how many schemas the model tends to induce.
% 
% ^.

The Guided Schema Induction, Fixed Attentions model extended the Unguided
Schema Induction, Fixed Attentions model by using RL to guide schema
induction. Whenever an exemplar was particularly useful (meaning it
reduced TD error by a thresholded amount), that exemplar was used
to induce a schema by comparing it to the candidate state. The threshold
used was in terms of standard deviations in reduction of TD error.
The reduction in TD error was computed by leaving out each exemplar
and computing TD without it. The difference between the TD error with
and without an exemplar is the reduction in TD for that exemplar.
Exemplars whose reductions were greater than 6 standard deviations
above the mean reduction were used to induce schemas. Thus in this
model, RL is used both to learn the values of the exemplars and to
guide schema induction.

The Guided Schema Induction, Learned Attentions model extended the
Guided Schema Induction, Fixed Attentions model by using RL to learn
exemplar-specific attention weights (the $u$ values). Each exemplar's
attention weight $u$ was initialized to 1. Over time the model uses
the TD error signal to update the attention weights so that particularly
useful exemplars have their attentions increased and misleading exemplars
have their attentions decreased (see Equation \ref{eq:deltaU}). Anytime
an exemplar's attention went below 0, it was pruned from the exemplar
pool. 

Each model variant was trained in blocks of 10 games of self-play.
In self play, at the start of each game the model variant was cloned
and played a game against itself. Learning (updates to $v$ and $u$)
was cached during the game and then applied after the game finished.
Learning occurred only during training. Following each block of 10
self-play games, the model was tested in a pair of games against an
ideal player (playing first in one game and second in the other).
The ideal player was given the correct values for every state and
always moved into the highest-valued next state. In testing games,
the model was given one point for each non-losing move it made (i.e.,
moves from which it could still guarantee a draw), for a maximum of
9 points per pair of testing games. In other words, points were awarded
for how long the model could play before the ideal opponent could
guarantee a win. More rigorously, the ideal player is defined by backward
induction, partitioning the state space into states from which ideal
play on both sides will lead to a draw, an X win, or an O win. Tic-tac-toe
has the property that the initial (blank) state lies in the first
of these three subsets. The model's score is determined by how long
it keeps the game in that regime.

\begin{figure}[ht]
\centering
\includegraphics[width=1\textwidth]{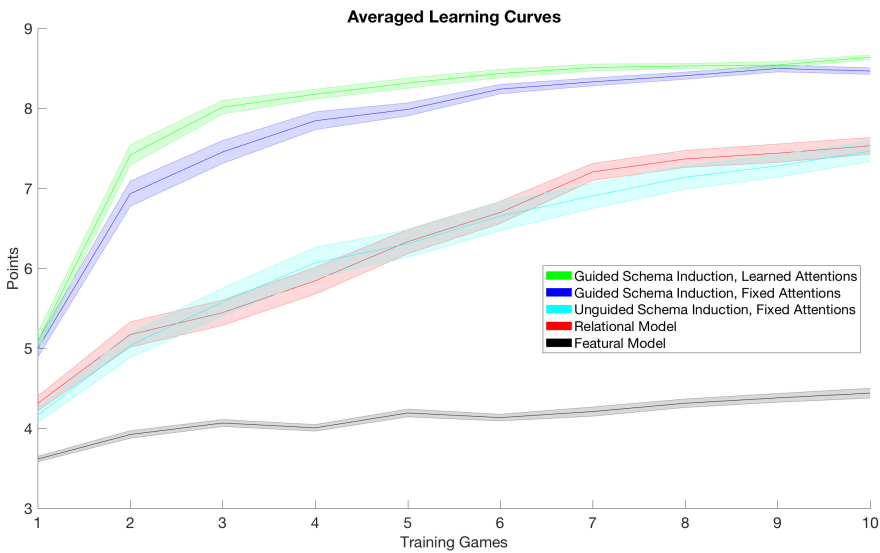}
\protect\caption{Simulation Results. Performance of the four model variations over 50,000 training games. The featural model (black) is extremely slow to learn. The relational model (red) uses analogical generalization and performs better than the featural model. Adding schema induction (light blue) to the relational model does not improve performance, unless the schema induction process is guided by reinforcement learning (dark blue). Adding the exemplar-specific attention learning mechanism (green) to the guided schema induction model further improves performance. Performance is measured by points, which is the number of moves made before the game is a sure loss against an ideal player, and is averaged across 64 independent copies of the model. Shading around each line indicates standard error bars. Training games are labeled in units of 5000 games.}
\label{fig:simulationResults}
\end{figure}

\section{Results}

Averaged learning curves are shown in Figure~\ref{fig:simulationResults}
for 64 independent copies of each model over 5000 blocks (50,000
training games). These results show that the featural model (black) is extremely slow to learn. The relational model (red) uses analogical generalization and performs better than the featural model. Adding schema induction (light blue) to the relational model does not improve performance, unless the schema induction process is itself guided by the prediction error signal (dark blue). There is a further improvement in performance when the model can also learn to adapt its attentions based on prediction error with the exemplar-specific attention learning mechanism (green).

\begin{figure}[!b]
\captionsetup[subfigure]{justification=centering}
\begin{subfigure}{.5\linewidth}
  \centering
  \includegraphics[width=1\linewidth]{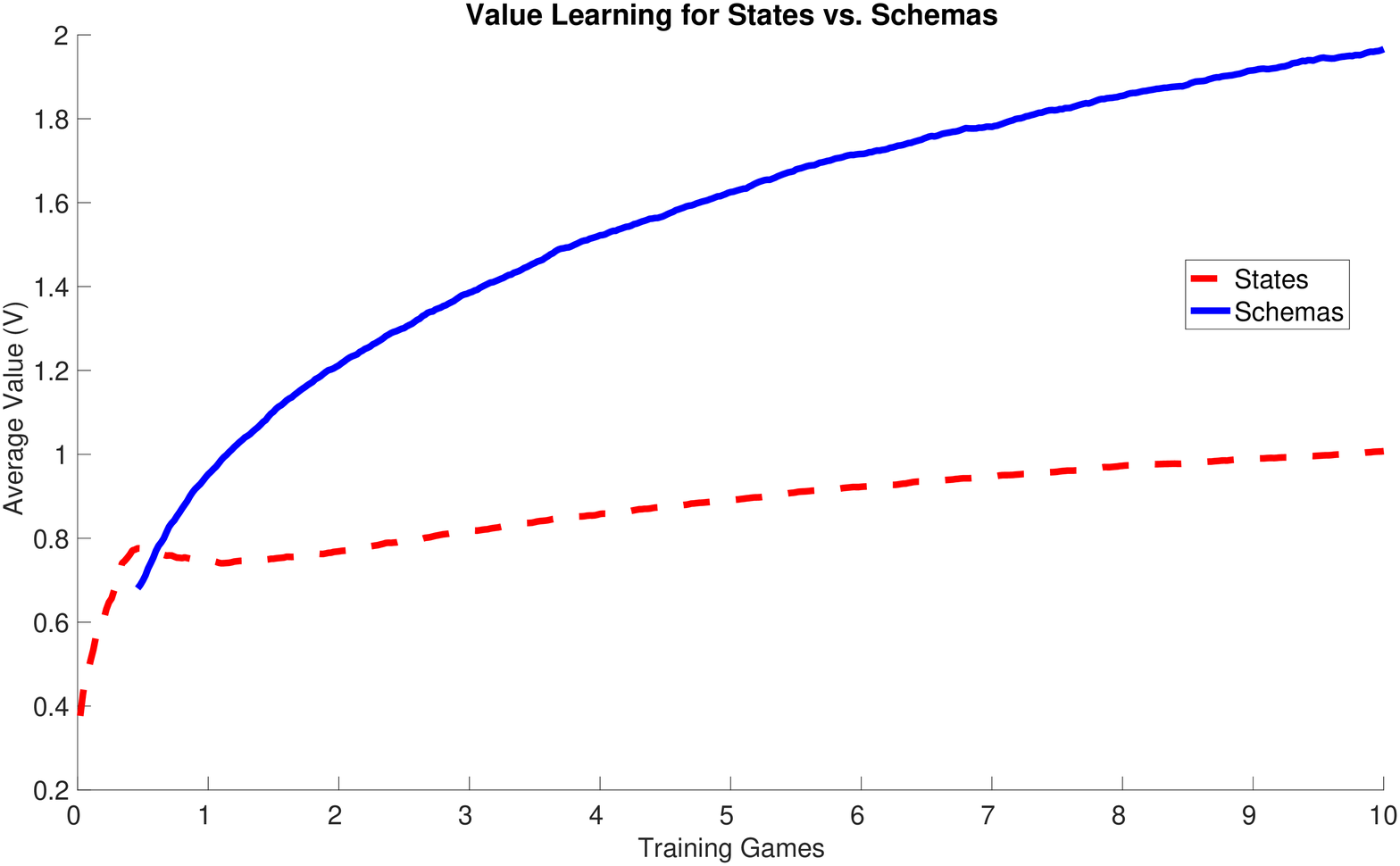}
  \caption{Value Learning}
  \label{fig:value_learning}
\end{subfigure}
\begin{subfigure}{.5\linewidth}
  \centering
  \includegraphics[width=1\linewidth]{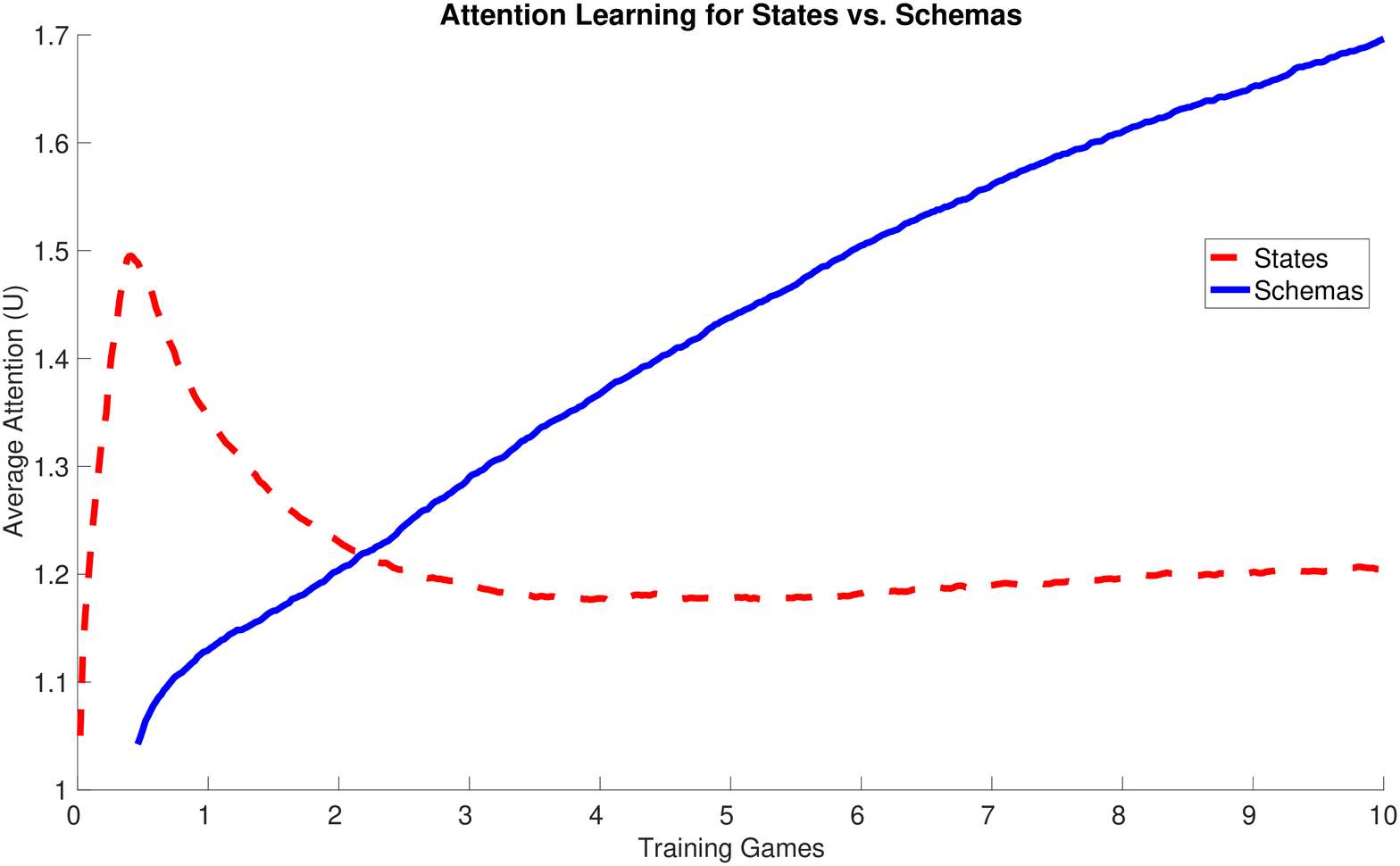}
  \caption{Attention Learning}
  \label{fig:attention_learning}
\end{subfigure}
\protect\caption{Value and Attention Learning for States vs. Schemas over 50,000 training games. Averages for states (red dotted lines) vs. schemas (solid blue lines) for 64 independent copies of the Guided Schema Induction, Learned Attentions model. (a) Average absolute value of \textit{V} for states vs. schemas as the model learns. Schemas quickly increase in absolute value compared to states, indicating that they are more diagnostic of winning and losing. (b) Once schemas begin to be induced, they rapidly rise in attention \textit{U} and competitively reduce attention to states, which indicates that the model is learning that schemas are more useful representations than states. Training games are labeled in units of 5000 games.}
\label{fig:value_and_attention_learning}
\end{figure}

As the model learns from training games, it updates its estimate of the value and attention for each exemplar. Figure \ref{fig:value_and_attention_learning} shows timelines of the model's values (\textit{v}) and attentions (\textit{u}) as the model learns from more training games. Values and attentions are plotted separately for states recruited as exemplars vs. schemas induced by the model. The values for schemas tend to be more extreme (either highly positive or highly negative) than the values for states, and so are more diagnostic of winning or losing board positions. Additionally, the attentions for schemas tend to be larger than attentions for states. An exemplar with a higher attention value tends to make especially accurate reward predictions. The overall larger attentions for schemas (as compared to states) indicates that schemas are, overall, more useful exemplars for reward prediction. The important conclusion is that the model learns to shift its representation of the environment from states that it has directly experienced to more abstract schematic representations.

\begin{figure}[ht]
\includegraphics[width=1\textwidth]{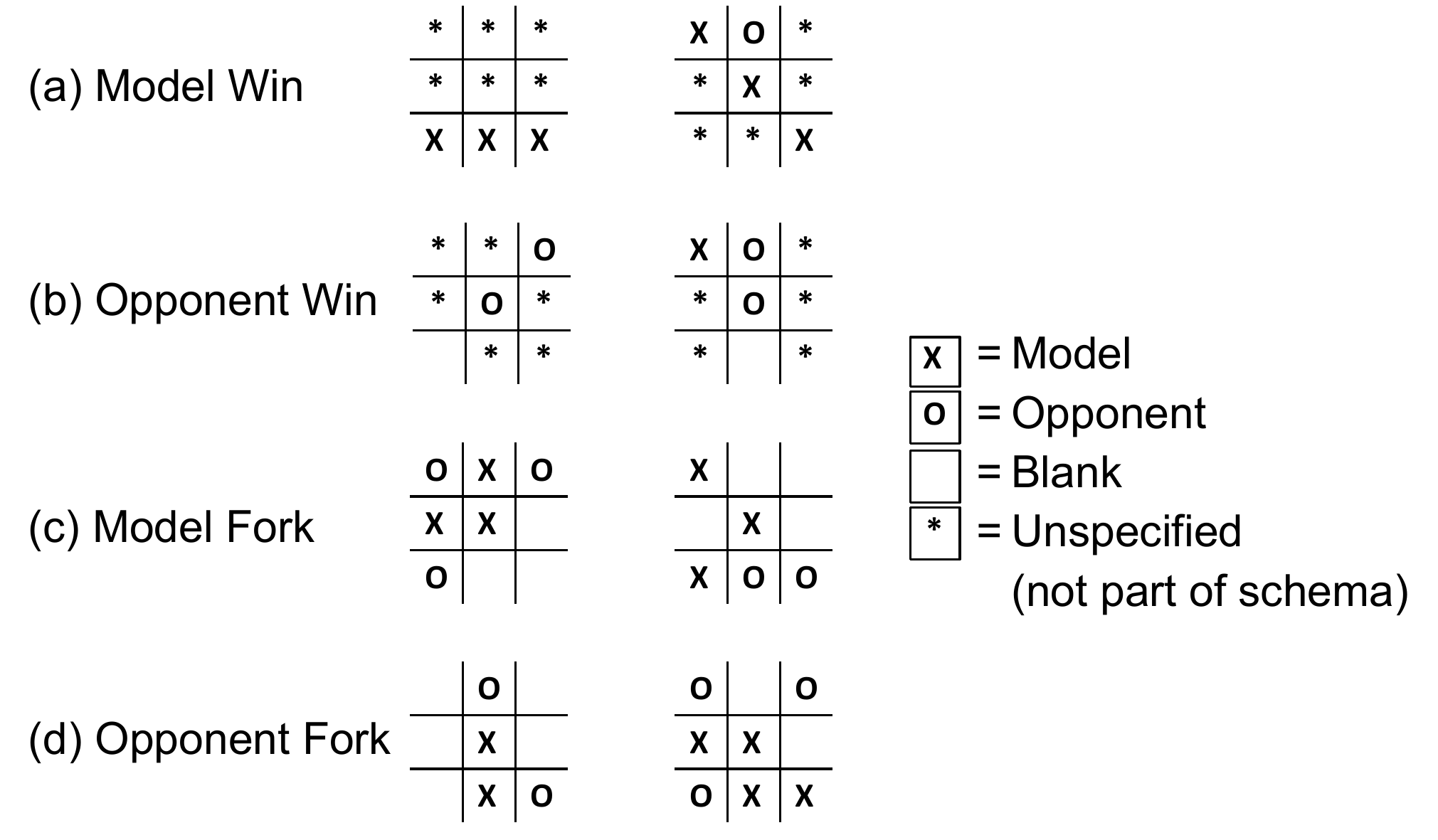}
\protect\caption{Examples of some of the representations with the highest attentions, as averaged over 64 independent copies of the Guided Schema Induction, Learned Attentions model after 50,000 training games. Because the model is implemented with afterstates, in each of these examples is valued from the perspective of the model (X) just having placed a token, and it being the opponent's (O) turn. The model associates (a) and (c) with positive reward value, and (b) and (d) with negative reward value, but all have high attentions because they are useful for predicting reward. (a) Examples of schemas where model wins the game. (b) Examples of schemas where the model is about to lose. (c) Examples of states where the model has two ways to win, and the opponent can only block one of them. (d) Examples of states where the opponent may have two ways to win. In the left board, a play by O in the top right corner will create two ways for the opponent to win. In the right board, the opponent is about to win, perhaps because it had produced a fork before X played in the left-center position.}
\label{fig:ticTacToeSchemas}
\end{figure}
%Instead, could show evolution of grids from early blocks to later blocks. 

Examples of some of the most
useful representations learned by the Schema Induction model are shown in 
Figure \ref{fig:ticTacToeSchemas}. The schemas in (a) represent winning states of tic-tac-toe.
The top left schema perfectly matches (with a similarity of 1) any
tic-tac-toe state with 3 X's in the bottom row, the left row, the right row, or the top
row - regardless of which objects (X, O, or blank)
occupy the other grid locations. Similarly, the left schema in (b) perfectly
matches any tic-tac-toe state with an O in the center, an O in a corner, and a blank in the opposite corner. This schema represents the state in which the
model is about to lose the game by an opponent making 3 O's along the diagonal.
These two schemas are ``pure'' or ``clean'' in that they contain no information that's
irrelevant to the winning or about-to-win states. However, the right schema in (a) does contain an irrelevant piece of information - the O in the top
center grid location. This schema is ``dirty'' in that it contains
irrelevant details from the states from which it was induced. With
further training, the model would likely refine this schema and learn
that abstracting out the O provides a more general, more useful schema.
The right schema in (b) is also "dirty" in that it contains the extraneous X in the top left corner.
The states in (c) represent something like "forks" in the tic-tac-toe domain, where the model has two ways to win and the opponent can only block one of them. The left state in (d) represents a state the model should avoid, because it provides opportunity for the opponent to create a fork by playing in the top right corner. In the right board in (d), the opponent is about to win, possibly because it had created a fork before X played in the left-center position.

\section{General Discussion}

The results presented here constitute a proof-of-principle that analogy
and schema induction can be productively integrated with a learning
framework founded on RL and similarity-based generalization. 
%\footnote{Analogical transfer is like similarity-based generalization, but it's also more sophisticated because it takes structure into account.} 
This integration leads to a model exhibiting sophisticated, abstract generalization
derived from analogical similarity, as well as discovery of new relational structures 
driven by their ability to predict reward.

The basic modeling framework used here applies not just to analogical
similarity and schema induction, but to other forms of representational
learning as well. Kernel-based RL offers a powerful and general theory
of representation learning, because it can be integrated with any
form of representation that yields a pairwise similarity function.
Its TD error signal can drive changes in representation via the objective
of improving generalization. This idea has been applied to learning
of selective attention among continuous stimulus dimensions \citep{jones2010}.
The current model offers a richer form of representation learning,
in that it acquires new concepts rather than reweighting existing
features.

% * <jmfoster@gmail.com> 2017-12-22T23:13:58.813Z:
% 
% Why so long to train?
% 1. Not the full model (lacking full schema induction & relational consolidation)
% 2. no look ahead / simulation
% 3. Nature of gradient descent operating in a large weight space
% 3. Explicit instruction to tic-tac-toe players
% 
% 
% ^.

\subsection{Related Models}

The analogical RL model also builds on other models of relational
learning. \citet{tomlinson2006pigeons} propose a model of analogical
category learning (BRIDGES), with essentially the same similarity
and exemplar generalization mechanisms adopted in the present model.
Our model adds to theirs in that it applies to dynamic tasks and in
that it grows its representation through schema induction. In BRIDGES,
analogy contributes to RL by providing relational generalization.
Our model also has the reverse, in that RL guides schema induction
and hence acquisition of abstract concepts. \citet{otterlo2012} has
developed methods for applying RL to relational representations of
the same sort used here, although the approach to learning is quite
different. His models are not psychologically motivated and hence
learn in batches and form massive conjunctive rules, with elaborate
updating schemes to keep track of the possible combinations of predicates.
In contrast, the present approach learns iteratively, behaves probabilistically,
and grows its representation more gradually and conservatively. This
approach is likely to provide a better account of human learning,
but a more interesting question may be whether it offers any performance
advantages from a pure machine-learning perspective.

In the present model, the activation of each exemplar elicited by
a candidate state can be thought of as a feature of that state. The
exemplar effectively has a ``receptive field'' within the state space,
defined by the similarity function. This duality between exemplar-
and feature-based representations is founded in the kernel framework
\citep[see][]{shawe-taylor2004}. The present model takes advantage
of this duality, producing a smooth transition from an episodic, similarity-based
representation to a more semantic, feature-based representation defined
by learned schemas.

The value and attention learning mechanisms have roots in traditional associational
learning models. The classic Rescorla-Wagner model introduced joint
learning of cue-outcome associations, but had no mechanism for attention
learning \citep{rescorla1972}. Although the Rescorla-Wagner model
allowed for different input cues to have different associabilities,
there was no mechanism for learning these attentions. \citet{mackintosh1975}
introduced an attention learning mechanism which modifies attention
to cues to reduce prediction error and reduce interference between
cues. Support for this attention learning mechanism comes from demonstrations
of learned inattention in experiments with rats \citep{mackintosh1971}
and humans \citep{kruschke2000}.

Although the analogical RL attention-learning mechanism in the present
work is being implemented in an exemplar model, it is compatible with
Mackintosh's theory and with extensions by \citet{kruschke2001}.
Each of these models would predict that attention would increase to
lower-variance cues  based on the idea that attention increases to
cues that are more predictive (contribute less error), compared to 
the average individual cue predictiveness (as in \citet{mackintosh1975}
and \citet{kruschke2001}'s mixture of experts model) or 
the overall combined cue predictiveness (as in \citet{kruschke2001}'s
EXIT model and the present ARL model). In the exemplar setting, the
model needs to learn which exemplars to retain in order
to interpolate the reward value of each new stimulus based on learned
exemplars. Thus the present ARL model also includes a normalization
in the definition of exemplar activation, which makes attention more
like a voting weight and less like salience. 

The present work is complementary to hierarchical Bayesian models
that discover relational structure through probabilistic inference
\citep{tenenbaum2011}. Whereas our model builds up schemas from simpler
representations, the Bayesian approach takes a top-down approach,
defining the complete space of possibilities a priori and then selecting
among them. The top-down approach applies to any learning model, because
any well-defined algorithm can always be circumscribed in terms of
its set of reachable knowledge states. This is a useful exercise for
identifying inductive biases and absolute limits of learning, but
it offers little insight into the constructive processes that actually
produce the learning. The present model offers proposals about the
mechanisms underlying how the human mind discovers new, abstract concepts. Extensions to this model are discussed in the following sections.

\subsection{Afterstate vs. Forestate Learning}

Although the tic-tac-toe domain is well captured with an afterstate
representation, most tasks can't use this simplification. In the forestate
version of the model, each state would have a set of available actions,
and the model would learn values for state-action pairings rather
than learning values just for (after)states.

Furthermore, the afterstate formulation misses an important way in which analogical transfer is more complex than similarity-based generalization (SBG), which doesn't accommodate the translation provided by the analogical mapping. In the extended forestate version of the present model, generalization is not just via blind similarity, because it takes the mapping (and hence the structure of the stimuli) into account. In a sense, the action is \textit{in} the schema. A forestate model wherein structure mapping informed only the similarity computation and not the mapping of actions would fail in cases where the extended model succeeds. 
%Consider an example from \citet{markman1993}, in which a pair of images depict analogous causal scenes. In the first image, a woman is receiving food from a community food bank delivery driver. In the second scene, the woman is giving food away to a squirrel. SBG would infer that "the woman is receiving food" applies to the second picture because it applies to the first and because the two scenes are similar [see also][]\citep{jones2006}. 
Consider the fork example from chess in Figure \ref{fig:chess-1}. SBG would infer that a capturing action involving the knight in the game on the left would apply to a knight in the game on the right. If instead the action is in the schema, then a capturing action by the knight on the left would be correctly translated through the mapping into a capturing action by the analogous queen on the right. 

% 
% the systematicity mechanism isn't used in the ttt simulation, because there are no hors. we could drop it entirely, or mention it in the gd as an extension (on par with learning state-action values), or keep it here and add an example of when and how it could matter.
% 

\subsection{Two-Stage Memory Retrieval}

A challenge for the present model is tractability, because it's not feasible to compute analogical mapping to all stored exemplars in memory. For computational efficiency and simplicity, the current implementation in the tic-tac-toe domain exploits knowledge of the game's invariance under a predetermined set of simple symmetries. A solution to the tractability problem for the full model that uses structural mapping is to incorporate two-stage memory retrieval, following the MAC/FAC model \citep{Forbus1995}. The first stage uses fast feature-vector similarity to efficiently retrieve a set of candidate exemplars, and the second stage uses structural alignment to determine the best analogical matches. Such two-stage retrieval enables a more computationally tractable and psychologically plausible form of analogical inference from stored exemplars to novel situations. 

The first stage of retrieval (Many Are Called) computes a MAC score which is used to select the set of candidate exemplars that pass on to the second (Few Are Called) stage. The candidate exemplar set is defined by the top N exemplars sorted by descending MAC score. To compute each exemplar's MAC score, first the model computes the cosine similarity between each exemplar's feature vector and the candidate state's feature vector. The choice of cosine similarity is an implementational detail, and the model is not theoretically committed to this particular choice of featural similarity. The present model's attention-learning mechanism can be incorporated into the MAC score by multiplying the featural similarity measure by the exemplar-specific attention weights u to compute each exemplar's MAC score:
\noindent 
\begin{equation}
\label{eq:macfac}
MAC(S,E)=u(E)^{p}\frac{f(S)\cdot f(E)}{\Vert f(S)\Vert\:\Vert f(E)\Vert}.
\end{equation} 
These two components (featural similarity and attention) in the MAC score make exemplars with higher featural similarity and higher attentions more likely to be retrieved from memory and included in the set of candidate exemplars. Thus more useful exemplars are more likely to be retrieved and relied upon for generalization. 
\footnote{In the existing model, exemplar attention \textit{u} is standing in for the expected value of retrieval, whereas in the extended model with two-stage retrieval, retrieval probability is linearly related to \textit{u}. Our theoretical commitment is that an exemplar with a higher \textit{u} has higher probability of being retrieved (monotonic relationship).}  
The relative weighting of the featural similarity and the exemplar attentions is set by an exponent p on the attention weight. The second stage of retrieval (Few Are Called) computes a FAC score through a more computationally complex structure-mapping process on relational representations. In the present model, the FAC score is the analogical similarity of Equation \ref{eq:analogical_similarity}. Although the mapping and similarity scoring process itself is not currently learned, it may be productive to enable the prediction error signal to influence mapping or scoring itself, as in \citet{liang2015learning}.

The MAC/FAC extension also lays groundwork that will be useful for a later model with relational consolidation, which is discussed in detail in the next section. In brief, the exemplars that are learned to be most useful (via the attention-learning mechanism described above) would become new perceptual features which could be leveraged by the MAC retrieval stage to improve the candidate exemplars retrieved for the subsequent structural alignment processing. 
%might move this to relational consolidation section
%
%I do have some preliminary results [see Masters Thesis]
%Want to extend to new domain. e.g., language, go  ?> make this my dissertation

\subsection{Relational Consolidation}

%for this paper, i suggest setting up consolidation as a next step in the development of the current model.  otherwise the connection to the rest of the paper seems weak.  

%At this point we have a model that can acquire new relational concepts, but there's a ceiling to what it can learn because all it can do is build configurations of the primitive elements it's endowed with. The metaphor is a one-story building. %In addition to explaining and arguing for consolidation, we could elaborate on how it would be implemented in our model.  this is all cast as a plan for future work, and we're excited to see what more this planned next generation of the model can do.

Although the present integration of analogy, schema induction, and
reinforcement learning proves powerful, there's a ceiling to what it can learn
because all it can do is build configurations of the primitive elements it's endowed with. 
It lacks a mechanism to create rich compositional hierarchies of relational concepts. 
Examples of such compositional hierarchies include computer architecture, mathematical functions, and natural languages, which 
all exemplify multiple levels of abstraction by chunking systems of
relations at one level into building blocks at the next level. In
computer architecture, digital logic gates are composed to form adders,
which are composed with other digital circuits to form an arithmetic
logic unit (ALU), which is a building block in a computer\textquoteright s
CPU. Software design manages complexity by continuing this hierarchy,
composing primitive functions into more complex functions, and from
there to objects and design patterns. The conceptual progression in
mathematics proceeds similarly, composing the counting operation to
define adding, which is further composed to form multiplying, and
then exponentiation. In traditional views of linguistics, phonemes,
morphemes, words, and sentences form another example of a relational
hierarchy.\footnote{Relational hierarchies are not taxonomic hierarchies. In a taxonomic
hierarchy, each concept or category is a union of lower-level categories.
In a relational hierarchy, each instance of a concept is a configuration
of instances of lower-order concepts.}

Although we agree with theories of schema induction, we argue it is
insufficient to explain human relational learning. Schemas are explicit
relational structures, and thus they cannot be bound to roles of yet-higher-order
relations in the way unitary objects and relations can. Experiments with chimpanzees suggest that newly learned relations
can only fill roles of other relations if they can be represented
as atomic entities \citep{Thompson1997}. Therefore, to explain acquisition of relational
hierarchies, we put forward the hypothesis that useful schemas are
eventually replaced (or supplemented) with unitary representations
\citep{foster2012}. Thus, a concept that was represented as a system
of relations (via the schema) can now be represented as an atomic
entity, capable of entering into relations itself. We label this process
relational consolidation, in a deliberate parallel to theories of
episodic memory consolidation \citep [e.g.,][]{squire1995}.

As summarized in Table \ref{table:ConsolidationConsequences},
consolidation is hypothesized to confer properties to a concept that
are not true of (unconsolidated) schemas, because consolidated concepts
are recognizable perceptually, without explicit (working-memory dependent)
structure mapping \citep{corral2014effects,corral2017learning,foster2012}. A consolidated concept can be recognized automatically, retrieved
from memory in parallel, and represented as an element of yet-higher-order
relations. 

\begin{table}[bt]
\caption{Predicted consequences of consolidation.}
\label{table:ConsolidationConsequences}
\begin{threeparttable}
\begin{tabular}{p{6cm} p{6cm}}
\headrow
\thead{Not Consolidated} & \thead{Consolidated}\\
More affected by WM demands & Less affected by WM demands\\
Quicker at analogical inference, because structure mapping is active & Easier to learn higher-order structure, because instances can be represented by tokens\\
Serial retrieval & Parallel retrieval\\
\hiderowcolors
\hline  % Please only put a hline at the end of the table
\end{tabular}
%\begin{tablenotes}
%\item JKL, just keep laughing; MN, merry noise.
%\end{tablenotes}
\end{threeparttable}
\end{table}

It is important to note that consolidation is not a change in the
declarative knowledge embodied by a concept. Rather, it is a proceduralization
of the concept that enables future changes in knowledge \textendash{}
similar to the interaction between declarative and procedural knowledge
in production systems \citep{anderson1998}.

%The DORA model of relational predication \citep{Doumas2008} has
%an operation similar to relational consolidation, in which it recruits
%a new proposition node to bind lower-order relations. This new node
%can be bound to roles of yet-higher-order relations, but the relation
%is still explicitly structured. In contrast, consolidation might be
%viewed within the DORA framework as enabling the new proposition node
%at the top of the relational structure to evolve into a new semantic
%node at the bottom. We conjecture this difference has important implications
%for recognition and retrieval of instances of the concept. 

The MAC/FAC model discussed above embodies the assumption that verifying the lower-level elements of an episode (i.e., predefined objects and relations) is fast and automatic, whereas verifying relational structure is slower and requires working-memory resources \citep{Forbus1995}. From this perspective, relational consolidation enables higher-order relational structure to be chunked and treated as a dimension of the feature vector used for memory probing. Prior to consolidation, retrieval of instances of a higher-order relational structure requires something
like the FAC stage, in which agents explicitly map between those
instances and the schema. Following consolidation, retrieval can rely
solely on the MAC stage, thus operating much more rapidly and without
requiring working memory. We also propose a similar difference for
perceptual recognition of instances of the concept. Before consolidation,
episodes must be structurally aligned to a schema. After consolidation,
an instance of the concept is explicitly represented and bound to
the lower-order relations. 

%
%In the first stage of analogical retrieval (Many Are Called),
%the target episode is converted to a flat feature vector that is used
%to probe all episodic memories in parallel. Importantly, the dimensions
%in the MAC feature vector are predefined, based on the concepts the
%learner currently knows. Stored episodes that share content (objects
%and relations) with the target are retrieved, without regard for how
%those objects and relations are connected by role binding. In the
%second stage (Few Are Chosen), the episodes retrieved by the MAC stage
%are filtered by structural alignment to the target. Only those episodes
%that are alignable with the target survive this stage. 
%%don't be redundant with MAC/FAC section above

We further propose that analogy, schema induction, and relational
consolidation form a cycle that, when iterated, can produce relational
hierarchies of arbitrary depth (height). This form of learning leads
to a dualist view of objects and relations, in which (nearly) every
concept is both a relational structure among its components and an
object capable of participating in relations. The conceptual systems
built from this hierarchical relational chunking are potentially quite
powerful and flexible.
%,... enabling humans to compose one-story conceptual buildings into skyscrapers.

%  The following is more appropriate as evidence for the attention-learning mechanism, not for consolidation
%We have hypothesized that factors determining which exemplars become consolidated may include frequency of instantiation and reward predictiveness. Our experiments with humans have shown that each of these factors does influence the weighting of exemplars when people are faced with ambiguous stimuli [cite Shapes and Oysters experiments]. %However, we have not yet found evidence of consolidation, but are excited to try.
% 
%Working on verifying these mechanisms with human experimental data (cite master's thesis)
%Tests of model?s schema strength mechanism
%Shapes: Higher frequency does seem to bias the retrieval and application of schemas, perhaps by increasing their strength or baseline activation
%Oysters: Participants appeared to weight the lower-variance category more in estimating the value of the combined cue stimulus. 

\subsection{Conclusions}

This paper has proposed a psychological theory that integrates reinforcement learning with relational representations and analogy. The integration produces a computational synergy in which analogy enables abstract generalization, and reinforcement learning drives discovery of useful relational concepts without relying on hand-coded representations. Analogy contributes mechanisms to generalize based on relational similarity, translate rewarding actions between mapped scenarios via analogical transfer, and produce progressively more abstract representations of the environment via schema induction. In return, reinforcement learning contributes mechanisms to learn the long-term value of exemplars, guide schema induction, and learn exemplar-specific attentions based on reward predictiveness. These mutually supportive mechanisms combine in a way that begins to explain how people and machines can learn abstract concepts based on experience with the world.

%\section*{acknowledgements}
%Acknowledgements should include contributions from anyone who does not meet the criteria for authorship (for example, to recognize contributions from people who provided technical help, collation of data, writing assistance, acquisition of funding, or a department chairperson who provided general support), as well as any funding or other support information.

%\printendnotes

% Submissions are not required to reflect the precise reference formatting of the journal (use of italics, bold etc.), however it is important that all key elements of each reference are included.
\bibliography{Foster_Jones_2017}

\end{document}